\def\year{2021}\relax
\documentclass[letterpaper]{article} \usepackage{aaai21}  \usepackage{times}  \usepackage{helvet} \usepackage{courier}  \usepackage[hyphens]{url}  \usepackage{graphicx}    \usepackage{natbib}  \usepackage{caption}     

\makeatletter
\let\copyright@on\relax
\makeatother

\usepackage{amsmath,amsfonts,bm}

\DeclareMathOperator*{\argmin}{arg\,min}
 \usepackage[switch]{lineno}
\usepackage[dvipsnames]{xcolor}
\usepackage{booktabs}
\usepackage{algorithm}
\usepackage{algpseudocode}
\usepackage{subfigure}

\usepackage{tabularx}
\usepackage{makecell}
\usepackage[inline]{enumitem}
\usepackage{multirow}
\usepackage{xspace}

\newcount\aaaiyear
\aaaiyear=\year
\newcount\year
\year=\aaaiyear
\usepackage{glossaries}
\GlsSetQuote{+}
\glsdisablehyper
\makenoidxglossaries
\glsdisablehyper
\newacronym{PPS}{PPS}{Planning for Policy Search}
\newacronym{MPC}{MPC}{Model-Predictive-Control}
\newacronym{GPS}{GPS}{Guided Policy Search}
\newacronym{iLQR}{iLQR}{iterative Linear Quadratic Regulator}
\newacronym{DDPG}{DDPG}{Deep Deterministic Policy Gradient}
\newacronym{PPO}{PPO}{Proximal Policy Optimization}
\newacronym{AQR-RRT}{AQR-RRT}{Affine Quadratic Regulator-Rapidly exploring Random Trees}
\newacronym{D-RL}{D-RL}{Deep-Reinforcement Learning}
\newacronym{RRT}{RRT}{Rapidly Exploring Random Tree}
\newacronym{AQR}{AQR}{Affine Quadratic Regulator}
\newacronym{LQR}{LQR}{Linear Quadratic Regulator}
\newacronym{SAC}{SAC}{Soft Actor-Critic}
\newacronym{TRPO}{TRPO}{Trust-Region Policy Optimization}
\newacronym{FIFO}{FIFO}{First-In-First-Out}
\newacronym{MCTS}{MCTS}{Monte-Carlo Tree Search}
 
\makeatletter
\newcommand\notsotiny{\@setfontsize\notsotiny\@vipt\@viipt}
\makeatother

\newcommand{\mycite}[1]{\cite{#1}}

\definecolor{orchid}{rgb}{0.85, 0.44, 0.84}
\definecolor{palecopper}{rgb}{0.85, 0.54, 0.4}
\definecolor{sapgreen}{rgb}{0.31, 0.49, 0.16}
\definecolor{amber(sae/ece)}{rgb}{1.0, 0.49, 0.0}
\definecolor{mediumpersianblue}{rgb}{0.0, 0.4, 0.65}

\newcommand{\xtarget}{x_{r}}
\newcommand{\vclosest}{v}

\newcommand{\vnew}{v_{\text{new}}}
\newcommand{\snew}{s_{\text{new}}}
\newcommand{\dnew}{d_{\text{new}}}

\newcommand{\affnzd}{\mathfrak{a}}
 
\title{Improving the Exploration of Deep Reinforcement Learning in Continuous Domains using Planning for Policy Search}
\author{
Jakob J. Hollenstein,
  Erwan Renaudo,
  Matteo Saveriano,
  Justus Piater
\\
}
\affiliations{
\textsuperscript{\rm 1}Department of Computer Science, University of Innsbruck, Innsbruck, Austria\\
    \{name.surname\}@uibk.ac.at
}

\makeatletter
\begin{document}

\graphicspath{{figures/}{./poster/_images/}{./poster/figures/}}
\maketitle

\begin{abstract}
  Local policy search is performed by most Deep Reinforcement Learning
  (D-RL) methods, which increases the risk of getting trapped in a
  local minimum. Furthermore, the
  availability of a simulation model is not fully exploited in D-RL
  even in simulation-based training, which potentially decreases
  efficiency.
To better exploit simulation models in policy search, we propose to
  integrate a kinodynamic planner in the exploration strategy
  and to learn a control policy in an offline fashion from
  the generated environment interactions. We call the resulting
  model-based reinforcement learning method PPS (Planning for Policy
  Search). We compare PPS with state-of-the-art D-RL methods in
  typical RL settings including underactuated systems. The comparison
  shows that PPS, guided by the kinodynamic planner, collects data
  from a wider region of the state space. This generates training data
  that helps PPS discover better policies.

 \end{abstract}
\newcommand\teximage[1]{\input{#1}}
\newcommand\soutpars[1]{\let\helpcmd\sout\parhelp#1\par\relax\relax}
\long\def\parhelp#1\par#2\relax{\helpcmd{#1}\ifx\relax#2\else\par\parhelp#2\relax\fi }

\newcommand{\refalg}[1]{Algorithm~\ref{#1}\xspace}
\newcommand{\reffig}[1]{Figure~\ref{#1}\xspace}
\newcommand{\refsec}[1]{Sec.~\ref{#1}\xspace}
\newcommand{\reftbl}[1]{Table \ref{#1}\xspace}
\newcommand{\Exp}{\mathbb{E}}
\newcommand{\weirdD}{\mathcal{D}}
\newcommand{\vect}[1]{\boldsymbol{#1}}
\newcommand{\grad}{\nabla}
\definecolor{superlightgray}{rgb}{0.95, 0.95, 0.95}
\newcommand{\old}[1]{{\colorbox{superlightgray}{\transparent{0.3}\parbox{\columnwidth}{#1}}}}
\section{Introduction}\label{sec:intro}
Robots in human-centric environments are confronted with less
structured, more varied, and more quickly changing situations than in
typical automated manufacturing environments. Research in autonomous
robots addresses these challenges using trial-and-error based learning
methods. However, learning by trying out actions directly on a real
robot is time-consuming and potentially dangerous for the environment
and the robot itself. In contrast, physically-based simulation
provides the benefit of faster, cheaper, and safer ways for robot
learning.

In this context, \gls{D-RL} has shown promising
results \cite{openai2018learning}, but the training with \gls{D-RL}
algorithms can be tedious and it is often resource demanding.  One
problem in \gls{D-RL} training is that most algorithms are
gradient-based and thus susceptible to local optima. This is
usually countered by adding stochasticity to the action
selection \cite{lillicrapContinuousControlDeep2016,haarnojaSoftActorCriticAlgorithms2019,plappertParameterSpaceNoise2017}.
Although it is known that gradient-based algorithms suffer from the
local minima problem, only some
papers \cite{plappertParameterSpaceNoise2017,hendersonDeepReinforcementLearning2018} mention it, while the
practical implications of the local minima problem are not
exhaustively investigated in the \gls{D-RL} literature.
\begin{figure}[t]
  \centering{
    \resizebox{\columnwidth}{!}{
      \def\svgwidth{\columnwidth}
      \begin{notsotiny}
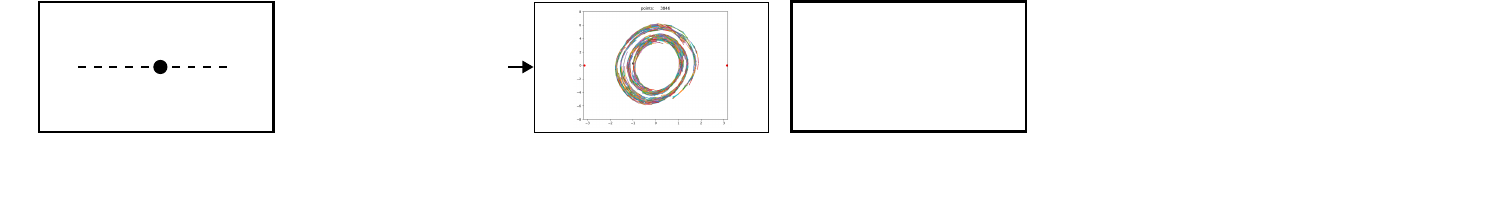
      \end{notsotiny}
    }}
  \centering{
    \resizebox{\columnwidth}{!}{
      \def\svgwidth{\columnwidth}
      \begin{notsotiny}
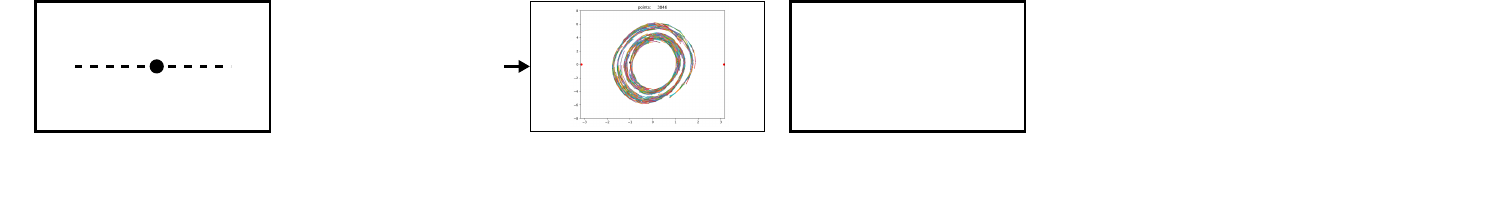
      \end{notsotiny}
    }}
\caption{A comparison between PPS and other D-RL methods. In PPS
    (top) a kinodynamic planner generates interactions---$(s,a,r,s')$
    tuples---that are used for offline policy search. In the typical D-RL
    setting (bottom) the agent directly learns a policy by performing
    exploratory actions to generate interactions.} \label{fig:datagen_both}
\end{figure}

We believe that insufficient exploration plays a major role in the
problem of learning sub-optimal policies. To remedy this problem, one
might increase the search time while keeping the exploration noise
high, or, as in this work, use more principled exploration.
In the limit increasing search time and exploring using randomly
sampled actions, would also yield acceptable solutions. However,
exploring by choosing actions to maximize exploration (\emph{directed
  exploration}) appears more promising to find good solutions more
reliably and in less time.
We focus on the latter
approach and build our method around two key considerations, namely
that \textit{i)}~covering a \emph{more diverse area} of the state
space increases the chances of finding an optimal solution, and that
\textit{ii)}~moving towards a directed exploration that exploits local
dynamics reduces the number of samples required to learn a good
policy.

Aiming for diverse and directed exploration and considering that
physically-based simulators are typically available for robotic
systems, we propose to exploit model-based and physically consistent
kinodynamic planners to generate environment interactions, thereby
\textit{guiding} the robot exploration, while tackling the problem of
planning time by synthesizing the planning results into a policy. We
select \gls{RRT}~\mycite{lavalleRapidlyExploringRandomTrees1998}
for kinodynamic
planning~\mycite{glassmanQuadraticRegulatorbasedHeuristic2010,
  perezLQRRRTOptimalSamplingbased2012a}, since this class of
model-based planning methods focuses on maximizing state-space
exploration (coverage). The data collected via \gls{RRT}-guided
exploration can then be used to learn a policy using an off-policy
search method. For policy search, the popular \gls{SAC}
approach~\mycite{haarnojaSoftActorCriticAlgorithms2019} has been
used. The way we used the robot model for the exploration makes the
proposed \emph{\gls{PPS}} approach a model-based reinforcement
learning
method~\mycite{suttonReinforcementLearningIntroduction2018}. In this
work, we present the PPS algorithm that combines planning and policy
search (see \reffig{fig:datagen_both}) and compare the performance of
PPS against the state-of-the-art D-RL approaches
\gls{DDPG}~\mycite{lillicrapContinuousControlDeep2016} and
\gls{SAC}~\mycite{haarnojaSoftActorCriticAlgorithms2019}. The aim of
this comparison is to answer the following research questions:

\begin{enumerate}[label=Q\arabic*]
\item \label{q_coverage}\label{q_traindata} How do the data generated by \gls{PPS}
  compare to those from other \gls{D-RL} methods -- do they cover a larger
  area of the state space?
\item \label{q_stuck} Are \gls{PPS} methods less susceptible to local
  optima than other \gls{D-RL} methods?
\end{enumerate}

Section~\ref{sec:related_work} presents an overview of the
related work. The \gls{PPS} approach is presented in
Sec.~\ref{sec:pps_method}. The experimental evaluation and the
comparison with existing approaches are carried out in
Sec.~\ref{sec:experiments}. Section~\ref{sec:discussion} discusses
obtained results, states the conclusions, and proposes further
extensions.

\section{Related Work}\label{sec:related_work}

Using physically-based simulations for learning is limited by the
necessity to approximate physical phenomena, causing discrepancies
between simulated and real-world results. This difference is called
the reality gap and it is a well-known problem in various fields of
robotics. For training in simulation and applying in the real world
(Sim2Real), filling this gap is crucial. One important method for
bridging this gap is Domain
randomization~\mycite{tobinDomainRandomizationTransferring2017a,
  sadeghiCAD2RLRealSingleImage2017,
  jamesTransferringEndtoEndVisuomotor2017}: instead of one simulated
environment, learning is done using a distribution of models with
varying properties such as mass, friction, and force/torque noise. The
idea is to make the behavior policies learned by the reinforcement
learning process more robust to the differences within this
distribution, thereby increasing robustness against the difference
between the training distribution and the target domain, i.e. against
the reality gap.
The work from OpenAI \mycite{openai2018learning} has shown a
successful use of domain randomization for learning
in-hand-manipulation; however, the number of required training steps is
raised by a factor of $33$ when domain randomizations are
introduced. This increases the number of training steps to the
magnitude of about $3.9\cdot 10^{10}$ from a magnitude of
$1.2 \cdot 10^9$, while classical deep reinforcement learning approaches
typically require $10^5$ to $10^9$ iterations of simulation
steps. Many algorithms are tested on $10^6$ timesteps,
depending on the environment. The required amount of training data can make this method
prohibitively expensive, and typically the availability of a simulation
model is not exploited.

\emph{Improving the efficiency} of domain randomization is an active
topic of research. Possible ways to increase sample efficiency include
using adversarial
randomizations~\mycite{mandlekarAdversariallyRobustPolicy2017},
limiting the training to stop before overfitting to idiosyncrasies of
the
simulation~\mycite{muratoreDomainRandomizationSimulationBased2018}, or
using model-based approaches for reinforcement
learning~\mycite{ChatzilygeroudisSurvey2020,SaverianoPIREM2017,levineGuidedPolicySearch2013}. In
the context of D-RL, the
\gls{GPS}~\mycite{levineGuidedPolicySearch2013} represents a prominent
model-based approach. In \gls{GPS}, rollouts from a deep neural
network controller are optimised by an optimal control method such as
the \gls{iLQR} \mycite{todorovGeneralizedIterativeLQG2005,
  tassaSynthesisStabilizationComplex2012} method. However, \gls{GPS}
is usually initialized from demonstrations since the exploration
capabilities of the underlying optimization method (\gls{iLQR}) are
limited. Furthermore, the optimization method requires a
specificaly-tailored cost function to guide the search procedure
towards relevant solutions.

The benefits arising from combining a model-based method with
model-free reinforcement learning were highlighted by
\citet{RenaudoGCK2014}. However, their work focuses on discrete
problems and their model-based and model-free algorithms
control the agent jointly, whereas we address continuous RL
problems where the planning method produces data for the policy
learner.

In this work, we identify \emph{insufficient exploration} as a major
cause of convergence towards \emph{suboptimal policies}, which results
in an increased training time to find a good policy. Therefore, we
propose a more principled exploration method guided by a kinodynamic
planner.

\gls{LQR}-\gls{RRT}~\mycite{perezLQRRRTOptimalSamplingbased2012a} and
\gls{AQR}-\gls{RRT}~\mycite{glassmanQuadraticRegulatorbasedHeuristic2010}
are examples of \gls{RRT} methods that use a dynamics-based cost
metric to guide the tree extension, enabling them to deal with
kinodynamic planning problems.

While \gls{RRT} methods are powerful asymptotically complete methods,
their computational cost is high, which only allows the plan to
be computed offline. Hence, RRT plans are not suitable to be directly
used as a control policy for continuous systems. The problem of
performance is also recognized in planning, and work is being
undertaken to make \gls{RRT} faster, for example
by~\citet{wolfslagRRTCoLearnKinodynamicPlanning2018}.

A possible alternative to \gls{RRT} could be \gls{MCTS}, which is part
of the successful AlphaGo~\mycite{silverMasteringGameGo2016}
system. However, vanilla \gls{MCTS} is not directly applicable to
continuous action domains.  A second advantage of \gls{RRT} over
\gls{MCTS} is that, by using local steering functions, they already
utilize the locally linearized dynamics often available in continuous
systems which can make the search more efficient. In \gls{PPS}, the
planner is used to explore the state space and to collect training
data from which a policy is learned. Hence, we do not need to plan at
run time when the robot executes the learning policy. We exploit
\gls{RRT} in the training phase as a reasonably powerful, yet
reasonably simple kinodynamic planner, although more complex
kinodynamic planners can also be used provided they collect
environment interactions.
In preliminary studies on the 1D Double Integrator benchmark domain
(see \reftbl{tbl:envdescription})
we found indications of increased exploration
\cite{hollensteinEvaluatingPlanningPolicy2019} in the data generated
using \gls{LQR}-\gls{RRT} as a kinodynamic planner compared to popular
\gls{D-RL} methods. We could already show that better exploration of
\gls{LQR}-\gls{RRT} generated data improved the performance of
policies learned from that data
\cite{hollensteinImprovingExplorationDeep2019,hollensteinHowDoesExplicit2020}.
In this work we extend on these results by moving from linearized to
affine-linearized dynamics, in both the distance metric, thus
replacing \gls{LQR}-\gls{RRT} with the \gls{AQR}-\gls{RRT}, as well as
the \gls{MPC} steering, and most importantly by applying the method to
benchmark domains with non-linear dynamics.
\section{Planning for Policy Search (PPS)}\label{sec:pps_method}
The key idea of PPS is to exploit kinodynamic planning to guide the
reinforcement learning exploration and collect interaction samples
$\{s,a,r,s'\}$ where $s$ and $s'$ represent the current and next state respectively,
$a$ is the performed action, and $r$ is the obtained reward. The
samples collected across different runs are stored in a replay buffer
$\mathcal{D}$ and are used to search for an optimal policy using any
off-policy search approach in an \emph{offline-} or
\emph{batch-}reinforcement learning fashion
\cite{ernstTreebasedBatchMode2005,langeBatchReinforcementLearning2012}.
The proposed PPS approach is summarized in Algorithm~\ref{alg:pps}. We
further use $x$ to denote hypothetical states used by the kinodynamic
planner which may or may not be feasible states of the environment,
$v$ denotes states stored as nodes in the internal tree of the
kinodynamic planner, and $d$ denotes trajectories. We also use affine
versions of all states $v^\affnzd, s^\affnzd, x^\affnzd$, which are
all related to the non-affine variants in the same way:
$x^\affnzd = \begin{bmatrix} x^T & 1 \end{bmatrix}^T$. Note that for
readability we will not explicitly include the conversions, but
instead assume both variants are available.

\subsection{Kinodynamic planning for interaction data collection}
We base PPS exploration on a variant of the Rapidly Exploring Random
Tree (RRT) that exploits locally affine dynamics and solves an Affine
Quadratic Regulator (AQR)
problem~\mycite{glassmanQuadraticRegulatorbasedHeuristic2010}. We
refer to this method as \gls{AQR}-\gls{RRT} in the rest of the
paper. During the exploration phase, the data are collected by
iteratively calling the \gls{AQR}-\gls{RRT} to extend the tree
$\mathcal{T}$
(\textsc{ExtendTree}$(\mathcal{T},
\xtarget)$ in Algorithm~\ref{alg:pps}).

An \gls{RRT} method consists of three components:
\begin{enumerate*}[label=\textit{\alph*)}]
\item a sampling method that decides where tree extensions should be directed to,
\item a distance metric that estimates the cost of going from nodes in the tree to a new target state, and
\item a local steering method, to reach from a given state to a target state.
\end{enumerate*}
\begin{algorithm}[t]
  \begin{algorithmic}
    \Function{PPS}{}
    \State $\mathcal{T} \gets \{v_0\}$\Comment{Initial State}
    \State $\mathcal{D} \gets \{\}$\Comment{Empty replay buffer}
    \For{$i \in \{1, \ldots, N\}$}
    \State  $\xtarget \gets \text{random state} $
    \State $\vnew, \dnew \gets $ \Call{ExtendTree}{$\mathcal{T}, \xtarget$} \Comment{See Alg.~\ref{alg:rrt}}
    \State $\mathcal{T} \gets \mathcal{T} \cup \{\vnew\} $ \Comment{Add node to tree}
    \State $\mathcal{D} \gets \mathcal{D} \cup \dnew$ \Comment{Add trajectory to $\mathcal{D}$}
    \EndFor
    \For{$i \in \{1, \ldots, M\}$}\Comment{Perform $M$ training steps}
    \State $\pi \gets$ \Call{OffPolicySearch}{$\mathcal{D}$} \Comment{See SAC}
    \EndFor
    \State \Return{$\pi$}\Comment{Found policy}
    \EndFunction
  \end{algorithmic}
  \caption{Planning for Policy Search (PPS)}
\label{alg:pps}
\end{algorithm}

In the current implementation of PPS, we start from an empty tree
$\mathcal{T}$ and add the initial robot state as the root node $v_0$.
Then, we loop $N$ times to incrementally extend the tree. At each
iteration, we uniformly sample the state space to create a new candidate
goal state $x_r$. The minimum distance from each node $v_i$ in the
tree to the new state $x_r$ is calculated using the \gls{AQR}
metric.
The metric uses $\dot{x} = Ax+Ba+c$ as its model, linearized around
$x_r$.
The affine term $c$ is incorporated into the affine matrices
$A^\affnzd =
\begin{bmatrix}
  A & c \\
  0 & 1
\end{bmatrix}$ and $B^\affnzd =
\begin{bmatrix}
  B^T &  0
\end{bmatrix}^T
$, which are used together with the affine state variants
($x^\affnzd, s^\affnzd, v^\affnzd$).

The bias term $c$, which is typically neglected in the standard
linearization, accounts for nonzero velocities at the
goal~\mycite{glassmanQuadraticRegulatorbasedHeuristic2010}.  This is
important since after extending the tree towards $x_r$, the reached
state will be used as the starting point for additional branches of
the tree, and reaching and staying (zero velocity) at $x_r$ would be
counterproductive. The complete calculation of the AQR metric is given
in the Appendix. The AQR distance to $x_r$, computed for each node in
the tree $\mathcal{T}$, is then used to determine the node
$\vclosest \in \mathcal{T}$ closest to $x_r$. This procedure to
determine $\vclosest$ is referred to as
$\text{AQR}_\text{Nearest}(\mathcal{T}, x_r)$ in
Algorithm~\ref{alg:rrt}.

In the steering method,
discrete-time linearized dynamics $G^\affnzd, H^\affnzd$ are used,
which are derived from the continuous-time linearized dynamics
$A^\affnzd, B^\affnzd$.
The node $\vclosest$ becomes a starting point, and a steering function
(\textsc{Steer}$(\vclosest, x_r, h, G^\affnzd, H^\affnzd)$ in Algorithm~\ref{alg:rrt}) is
applied to compute a feasible state trajectory that connects $\vclosest$
with $x_r$.

\begin{algorithm}[!tb]
\caption{Extend Tree with AQR-RRT}
  \begin{algorithmic}
    \Function{ExtendTree}{$\mathcal{T}, \xtarget$}
    \State  $h \gets \text{user defined horizon}$
    \State  $\vclosest \gets $\Call{$\text{AQR}_{\text{Nearest}}$}{$\mathcal{T}, \xtarget$}
    \State  $A^\affnzd, B^\affnzd \gets $\Call{env.dynamics}{$x_r$}
\State  $G^\affnzd, H^\affnzd \gets $\Call{discretize}{$A^\affnzd, B^\affnzd$}
\State $\snew, \dnew \gets $\Call{Steer}{$\vclosest, x_r, h, G^\affnzd, H^\affnzd$} \Comment{See Alg.~\ref{alg:steer}}
    \State $\text{where}$
    \State $~~\dnew = (s_0, r_0, a_0, s_1), \ldots, (s_{n-1}, r_{n-1}, a_{n-1}, s_{n})$
    \State $~~\snew = s_{n}$
\State $\vnew \gets $\Call{StorePoint}{$\snew, \dnew, \vclosest$}
    \State \Return $\vnew, \dnew$
\EndFunction
\end{algorithmic}
  \label{alg:rrt}
\end{algorithm}

We solve the steering problem using a linear
\gls{MPC} method \cite{camachoModelPredictiveControl2007}.
In the standard implementation, MPC works by repeatedly solving a
quadratic program (QP) from the current state $x_0 \gets \vclosest$ to the goal
$x_r$ over a finite time horizon $h$. The solution of the QP program
is a trajectory of control inputs $a_k,~k=0,\ldots,h-1$ that drives
the robot from $x_0$ to $x_r$ within $h$ steps. The first control
$a_0$ is then used to command the robot to attain a new state $s_1$.
This procedure (solving the QP and making one controlled step) is
repeated, where $x_0$ is replaced $x_0\gets s_1$. Starting from this
new state $x_0$ the QP is solved and another action $a_0$ is
taken. The system thus converges to the state $x_r$ and stays there.

In our implementation, we adopt a slightly modified version of the
linear MPC that makes use of a shrinking (in contrast to the receding)
horizon to let the robot reach the goal position with nonzero
velocity.  While we also start solving the QP with a horizon
$h_0 \gets h$ and take step $a_0$, in the next repetition, the horizon
$h_i$ is shortened by one step $h_{i+1}\gets h_{i}-1$. Thus we will
reach $x_r$ after a total of $h$ steps. If we did not shorten the
horizon and let the search run for $m$ steps, then we would expect to
reach the goal state $x_r$ at state $x_{h-m}$ of the current predicted
horizon; however, after this point $m$ more steps
($x_{h-m}, \ldots, x_m$) will be taken. Minimizing the cost associated
with these remaining steps drives the system away from $x_r$ if the
velocity at $x_r$ is nonzero, and thus motivates shortening the
horizon. In some cases, the environment's constraints might not allow
for exactly reaching $\xtarget$ and the optimization might become
infeasible and fail. In these cases, the state actually reached by the
steering procedure will be returned.
The complete steering method is given in the Appendix. The result of
the steering procedure is a new node $\vnew$ that is added to the tree
$\mathcal{T}$ and a trajectory $\dnew$ of intermediate states, actions,
and rewards that are added to the replay buffer $\mathcal{D}$.

\subsection{Transform interaction data into a control policy\label{sec:rrto2policy}}
As stated previously, the procedure used by \gls{RRT} to grow
the tree (see Algorithm~\ref{alg:rrt}) is computationally expensive
and becomes rapidly incompatible with real-time control requirements.
Learning a policy solves this issue since no planning step is needed
during the execution.  Importantly, we use the original \gls{RRT} and
not the \gls{RRT}$^*$\cite{karamanSamplingbasedAlgorithmsOptimal2011}
variant. The difference between \gls{RRT} and \gls{RRT}$^*$ is that
the latter iteratively reconnects and improves paths and thus
converges to shortest paths, which \gls{RRT} is not guaranteed to
do. However, this comes at a price: to reconnect nodes with
shorter paths, the steering function has to be used, which performs
environment steps. Thus the use of RRT$^*$ would incur additional
interactions on the simulated environment
to perform this optimization.
Since we first want to find areas close to the rewards and then, in
the future, perform local finetuning using the \gls{D-RL} method
directly, it does not make sense to spend environment steps on
improving all paths.

After the completion of the data collection, $\mathcal{D}$ consists of
a set of simulated (environment) interactions that can be used to search for an
optimal control policy using an off-policy search method. We collect
the data by maximizing the coverage of the state space, rather than
searching for high-reward regions.

In PPS therefore $\mathcal{D}$ constitutes a fixed replay buffer and is
used to train a policy (\textsc{OffPolicySearch}$(\mathcal{D})$ in
Algorithm~\ref{alg:pps}) without generating new environment data during
the policy search. The learned policy is then evaluated on the
environment. In this work we use
\gls{SAC}~\cite{haarnojaSoftActorCriticAlgorithms2019} as the
off-policy learning method. It uses a stochastic policy and an
actor-critic approach. It explores by sampling actions from the
stochastic policy, and adds an entropy term to the value-function loss
to encourage more exploratory behaviour of the policy; that is, high
entropy in the action selection is encouraged.

\begin{figure*}[h]
  \centering
  \subfigure[1D double integrator]{\raisebox{1cm}{\includegraphics[width=0.3\columnwidth]{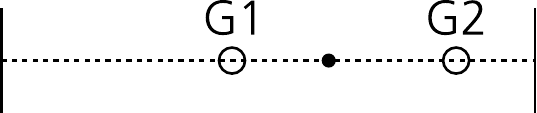}\hspace{2mm}
      \includegraphics[width=0.3\columnwidth]{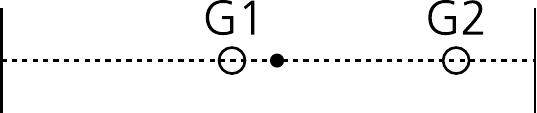}\hspace{2mm}
      \includegraphics[width=0.3\columnwidth]{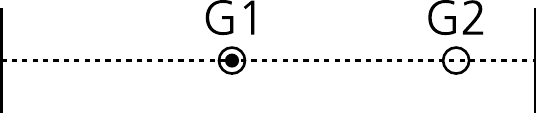}}}
  \hspace*{5mm}
  \subfigure[2-Link Planar Arm]{\includegraphics[width=0.3\columnwidth]{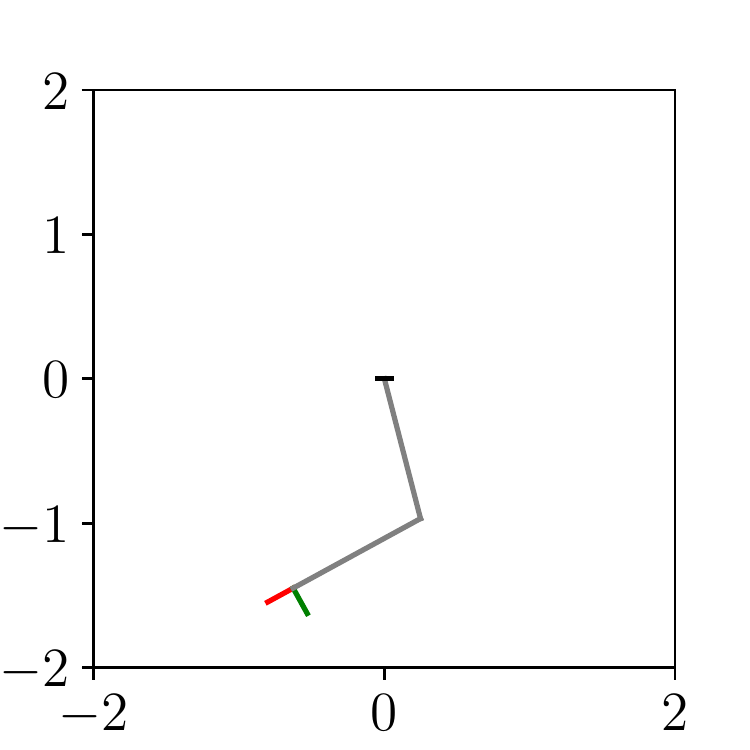}\hspace{2mm}
    \includegraphics[width=0.3\columnwidth]{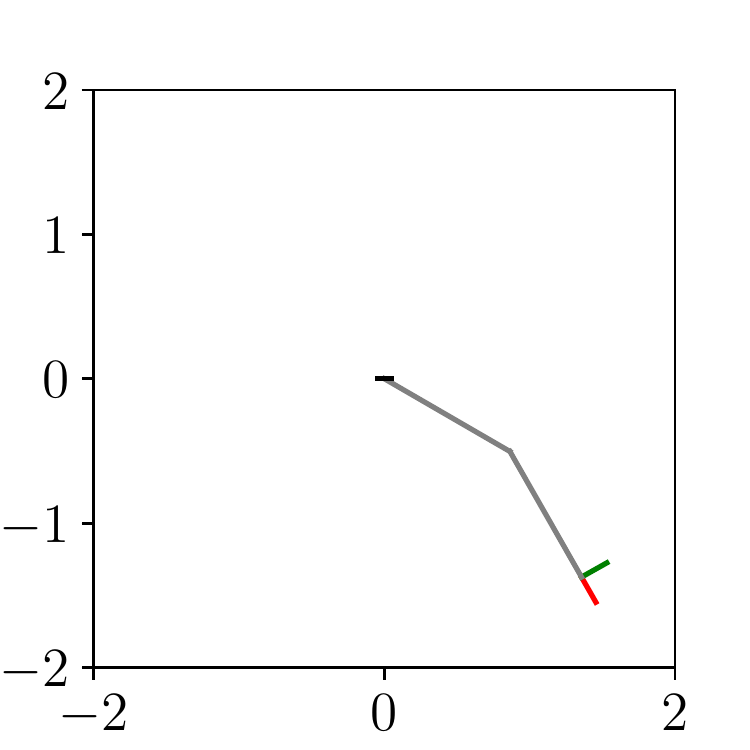}\hspace{2mm}
    \includegraphics[width=0.3\columnwidth]{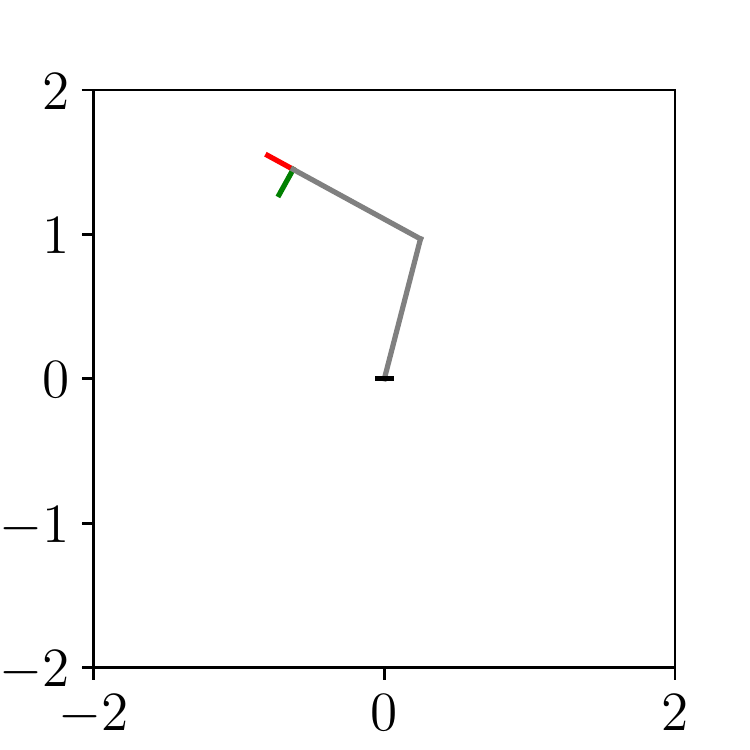}}

  \subfigure[Acrobot]{\includegraphics[width=0.3\columnwidth]{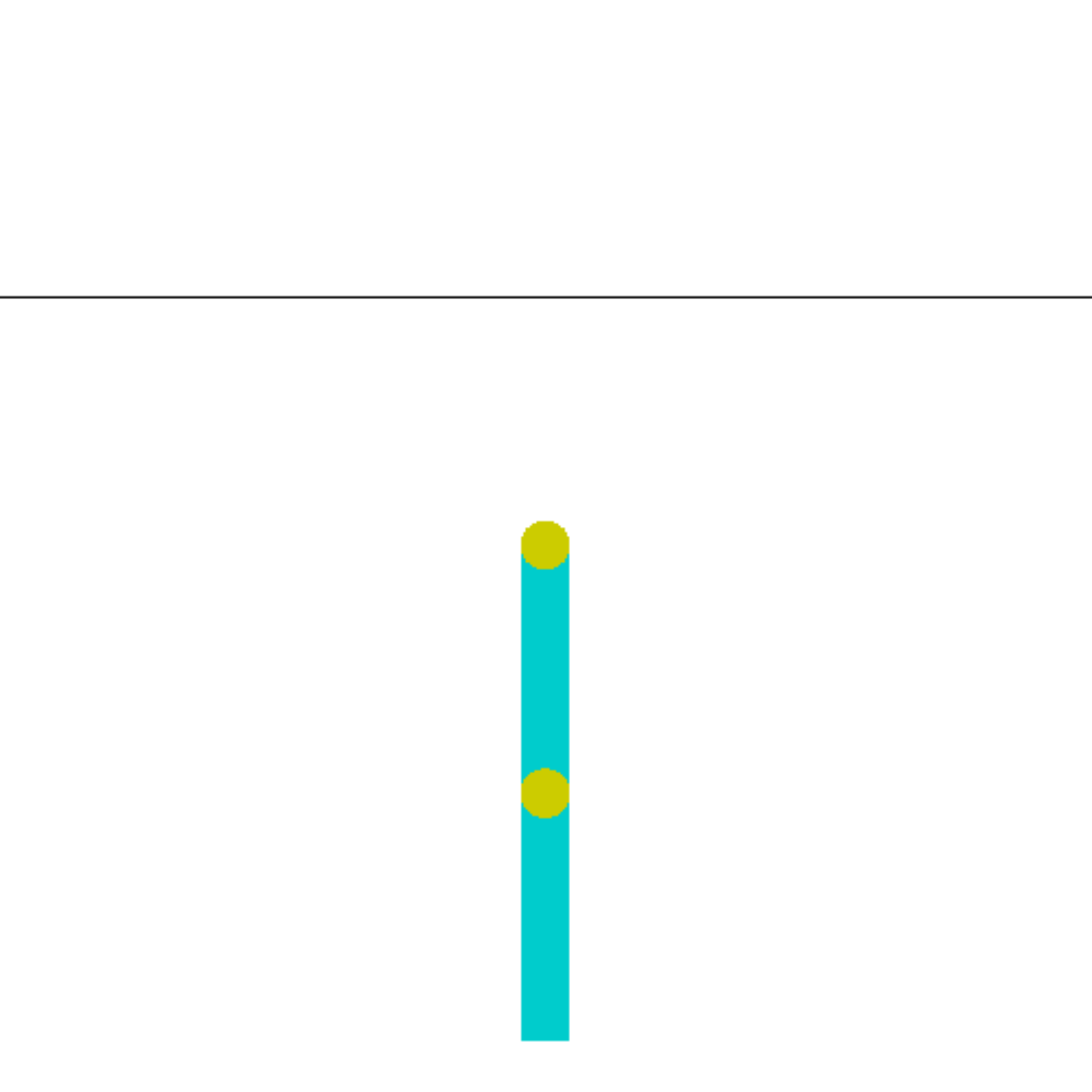}\hspace{2mm}
    \includegraphics[width=0.3\columnwidth]{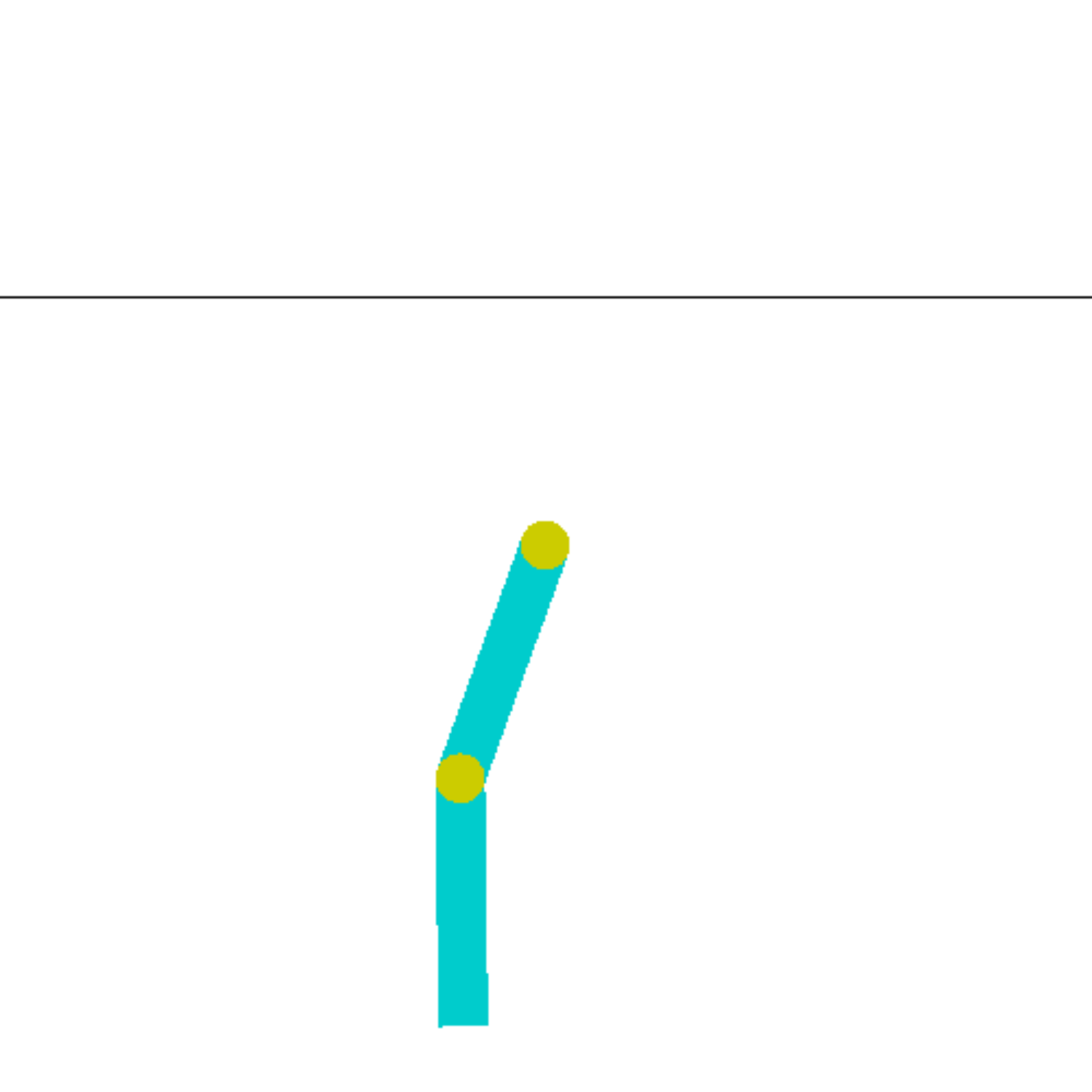}\hspace{2mm}
    \includegraphics[width=0.3\columnwidth]{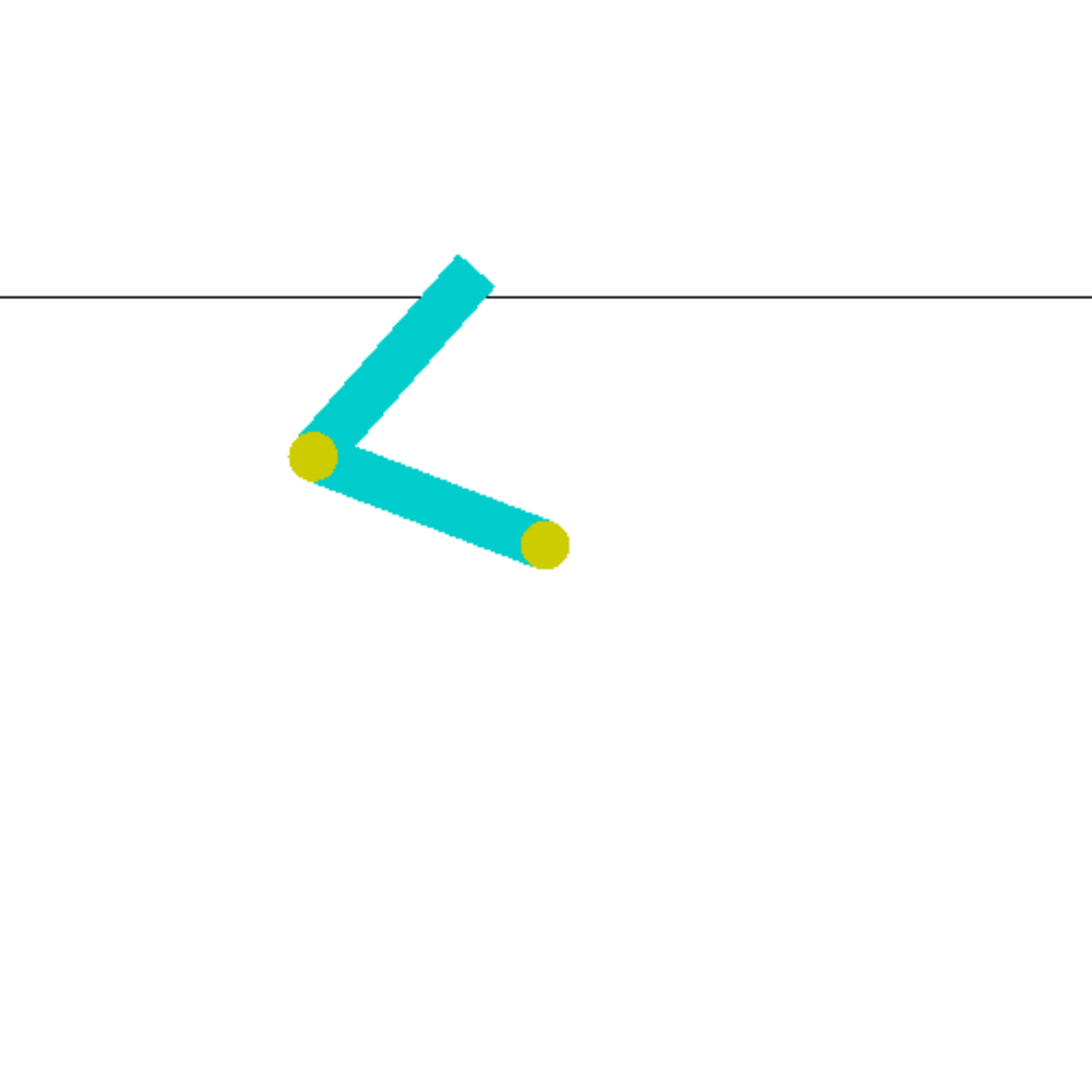}}
  \hspace*{5mm}
  \subfigure[Mountain car]{\includegraphics[width=0.3\columnwidth]{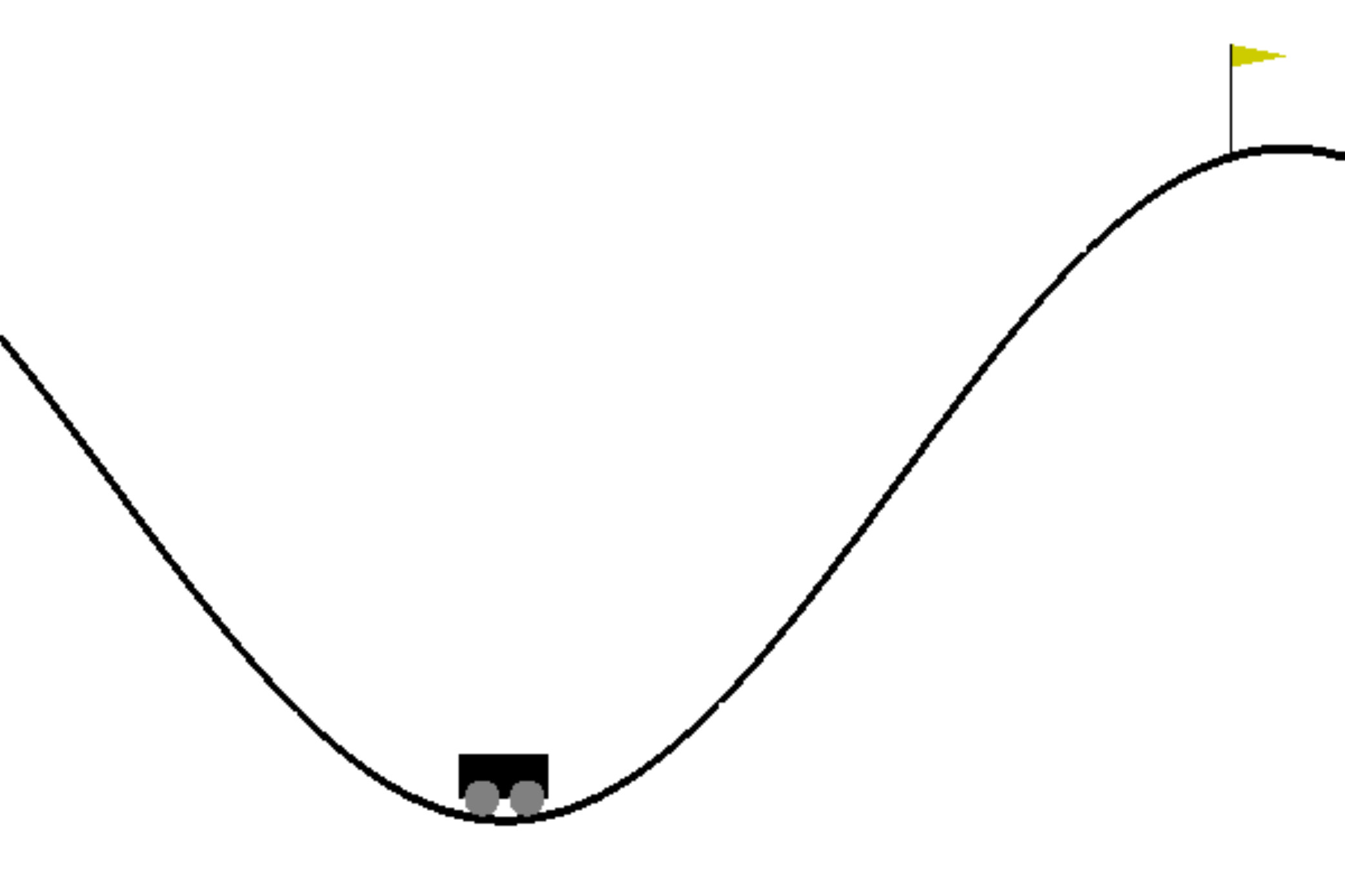}\hspace{2mm}
    \includegraphics[width=0.3\columnwidth]{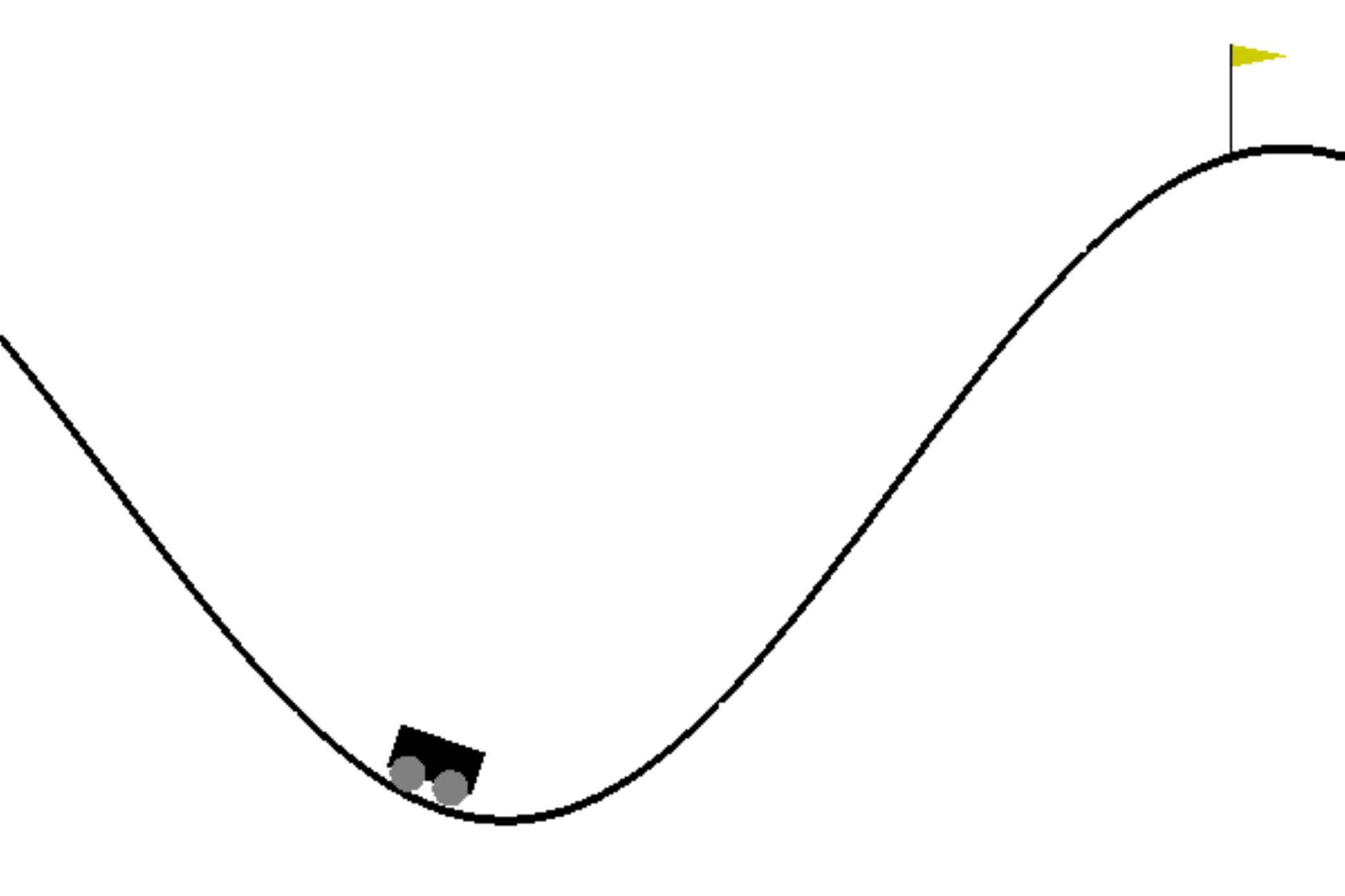}\hspace{2mm}
    \includegraphics[width=0.3\columnwidth]{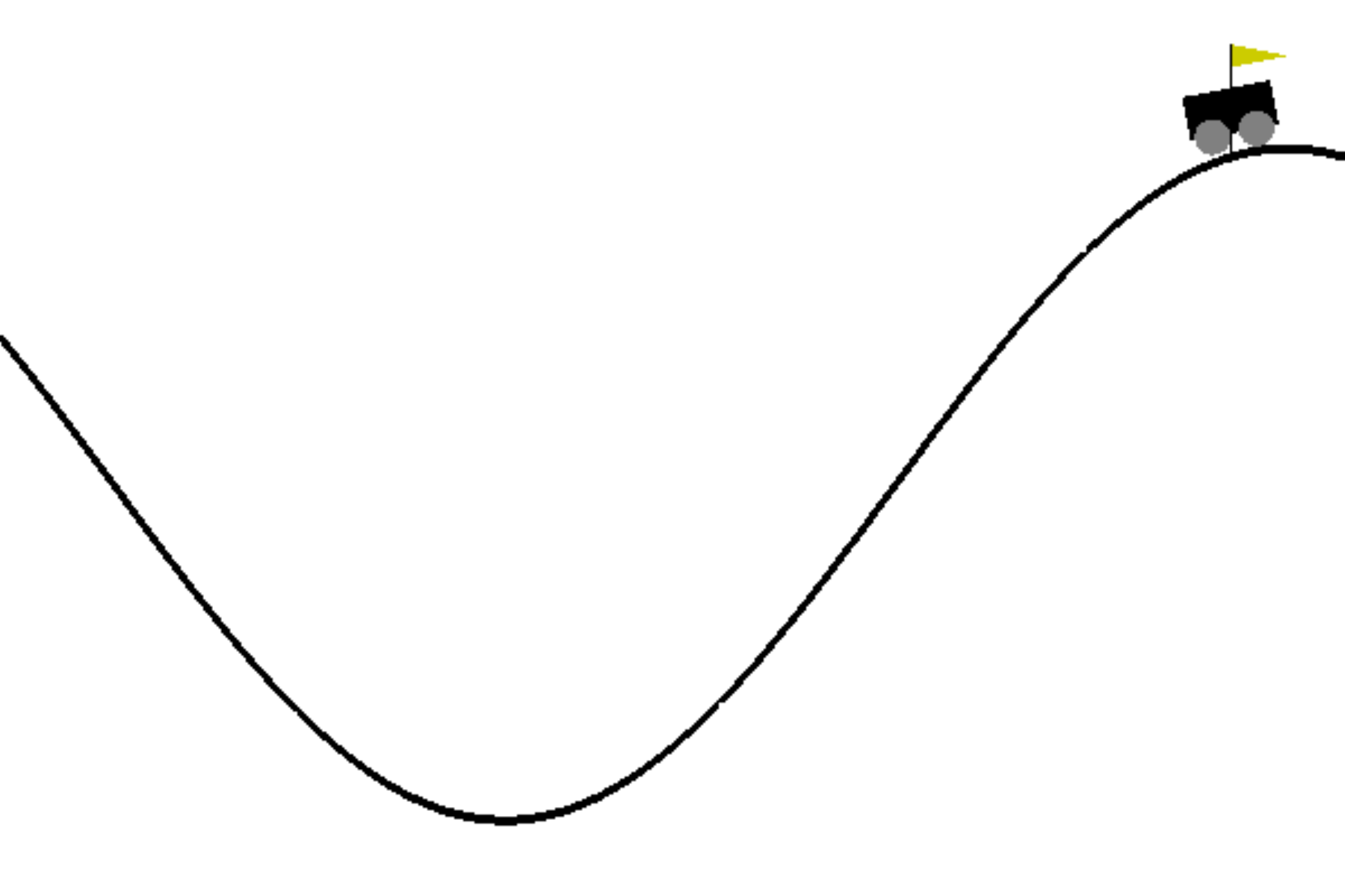}}

\caption{Environments used for testing. For each environment, the different snapshots show the initial, an intermediate, and the final condition.
  }
  \label{fig:environments}
\end{figure*}

\section{Experimental Evaluation}\label{sec:experiments}

Reinforcement learning agents collect experience and use that
experience to learn a policy, either implicitly in on-policy
algorithms such as
\gls{PPO} \cite{schulmanProximalPolicyOptimization2017} or
\gls{TRPO} \cite{schulmanTrustRegionPolicy2015}, or explicitly in
algorithms such as \gls{DDPG} or \gls{SAC} which use a replay
  buffer. Thus, the exploration process should reach regions in the
state space relevant for the task so that it can learn a
well-performing policy. If it cannot reach high-reward areas of the
state space, the learned policy will also not move the agent to these
regions and therefore the achieved return will be lower.

We assume that exploration and performance are linked and test
whether using the \gls{PPS} method increases exploration (\ref{q_coverage}).
We measure exploration by area of covered state space.

Since this is only a necessary but not a sufficient property, we also
test whether these data allow us, on average, to learn more succesful
policies, and thus are less susceptible to local
optima (\ref{q_stuck}). We measure this by learning policies from the
collected data and evaluating the returns of these policies.

We compare the performance to the popular off-policy \gls{D-RL}
algorithms \gls{DDPG} \cite{lillicrapContinuousControlDeep2016} and
\gls{SAC} (see Section~\ref{sec:rrto2policy}).

\gls{DDPG} is an off-policy method that learns a deterministic policy
using an actor-critic approach. Exploration is done by using the
deterministic policy and adding exploration noise to the selected
actions.

We use the implementations provided by \citet{stable-baselines} which
are a tuned and improved version of the algorithms provided by
\citet{baselines}.
We exclude the hyperparameters from the policy search and assume that
the algorithms are robust over a wide range of environments using the
default values of the hyperparameters.

\subsection{\ref{q_coverage} --  Comparing data generation}

To compare the exploration, we collect the data the agents see during
their learning phase (see \reffig{fig:datagen_both}). These data are
then analysed for state-space coverage. The coverage is calculated as
the percentage of nonempty bins.  For simplicity we use
uniformly-shaped bins. The number of bins is equal along each
state-space dimension and is set to
$\mathrm{divisions} = \left\lceil
  \left(\frac{N}{5}\right)^{\frac{1}{d}} \right\rceil$, i.e.\ such
that, in the uniform case, we expect five data points in each bin on
average. Because of the ceiling operator, the total number of bins
$n_{\text{bins}}^d = \mathrm{divisions}^d$ can exceed $N$.  This is
especially true in high dimensional cases, where this will lead to
many more bins than there are datapoints. To compensate for this, we
scale the resulting number of nonempty bins by
$ \max\left\{\frac{{n_{\text{bins}}}^d}{N}, 1\right\} $

\subsection{\ref{q_stuck} -- Susceptibility to local optima}

We perform training runs with \gls{DDPG}, \gls{SAC} and our \gls{PPS}
method. After the agents have learned, we use their policies to
generate evaluation returns, which we analyze to compare their
performance. This is done on $10$ independent learning runs.

While the \gls{DDPG} and \gls{SAC} agents learn directly on the
environment, our \gls{PPS} agent uses the planner
(\gls{AQR}-\gls{RRT}) to generate data. The generated data are stored
in a replay buffer. The replay buffer is fixed -- no experience is
added, no experience is removed and an off-policy algorithm
(\gls{SAC}) is used to learn from this buffer. As a baseline this
experiment is also performed for data generated by an \gls{SAC} agent
and by a \gls{DDPG} agent. The data of both are also used in fixed
\gls{SAC} replay buffers to learn policies. We fixed the \gls{SAC}
entropy coefficient hyperparameter to $0.1$\footnote{We tested entropy
  coefficients in the range of $0.005 \ldots 10$.} instead of
automatically adjusting it, since we noticed instability when learning
from a fixed buffer. To allow the same entropy coefficient across
different environments (reward ranges), we normalized the rewards in
$\mathcal{D}$ to have zero mean and unit standard deviation.

We compare the performance on four environments that are depicted in
\reffig{fig:environments}. In all four environments the agent starts
in the same state configuration and has to reach a goal region.

\begin{description}
\item[1D Double Integrator] is a force-controlled, one-di\-men\-sion\-al
  point mass in a position-wrap-around environment. The state space
  consists of position and velocity. The environment features two goal
  locations. The agent starts closer to a lower-reward goal region and
  has to move away from this region in order to reach the second,
  higher-reward goal location. For further details see \reftbl{tbl:envdescription} in the
  appendix.

\item[2-Link Planar Arm] is a torque-controlled, 2-link planar arm
  implemented following the description by \citet{zlajpahSimulationNrPlanar1998}. While the arm is torque
  controlled and the state space consists of joint positions and joint
  velocities, the reward is given in the task space.
  The direct path is blocked due to joint limits. The feasible path
  involves moving away from the task-space goal, incurring less
  reward, but ultimately achieving maximum returns.

\item[Acrobot] is the Acrobot \mycite{murrayCaseStudyApproximate1990}
  from OpenAI Gym \cite{brockmanOpenAIGym2016}.
We modified the agent to use continuous actions $a \in [-1, 1]$
    and changed the state representation to angle and angular
    velocity. In our version, the agent always starts in the
    lowest-energy position.
  \item[Mountain Car] is the continuous mountain
    car \cite{mooreEfficientMemorybasedLearning1990} from
    \citet{brockmanOpenAIGym2016}. Here we modified the starting
    position to be the lowest-energy state, which makes the
    environment harder, as it is less likely to be solved by chance.
\end{description}

For the latter three environments we used JAX \cite{jax2018github} to
derive the linearized dynamics and speed up computations.

  \begin{figure*}[t]
    \centering
    \includegraphics[width=\textwidth]{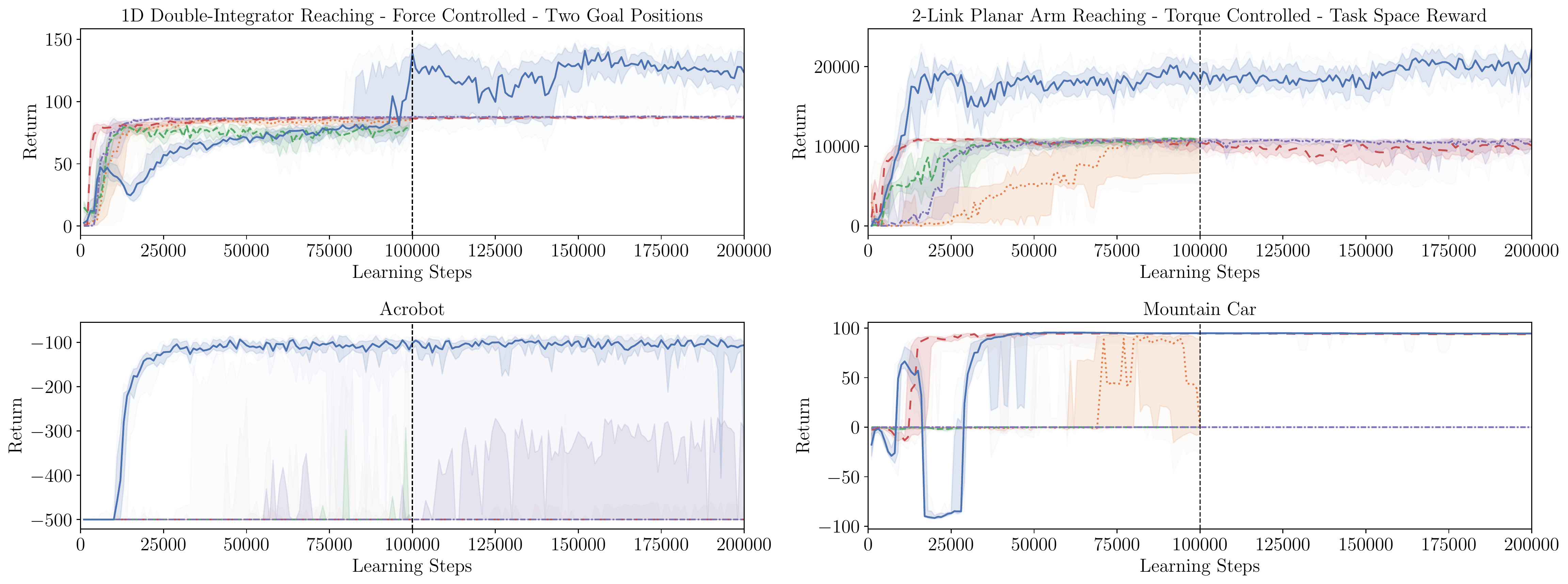}
    \includegraphics[width=\columnwidth]{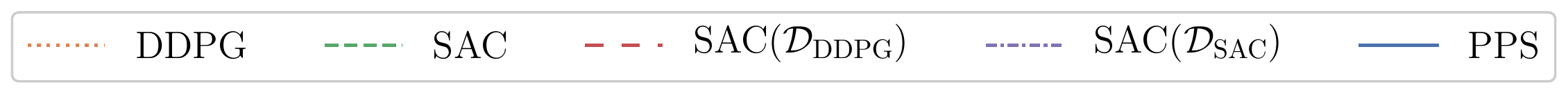}

\caption{Learning curves obtained with DDPG, SAC, and
      PPS (ours). The graphs show returns achieved by evaluation rollouts in the respective environment after training the policy for the given number of learning steps. The indirect methods ($\text{SAC}(\mathcal{D}_\text{DDPG}), \text{SAC}(\mathcal{D}_\text{SAC})$) and PPS are trained on a fixed replay buffer $\mathcal{D}$, filled with $10^5$ environment
      interactions (the $2\cdot10^5$ learning steps do not
      add any further interactions to $\mathcal{D}$).
DDPG and SAC, which directly learn as they collect environment interactions, are also shown for $10^5$ steps (and thus stop at the vertical center line).
      These $10^5$ collected environment interactions are used to train $\text{SAC}(\mathcal{D}_\text{DDPG})$ and
      $\text{SAC}(\mathcal{D}_\text{SAC})$.
}
  \label{fig:learning_curves}
  \end{figure*}
\section{Results}

\subsection{\ref{q_coverage} -- Comparing data generation}\label{a_coverage}
\begin{table}

  \renewcommand{\arraystretch}{2}

\begin{tabularx}{\columnwidth}{@{} l X X X @{}}
\toprule
     \makecell{Environment} &                \makecell{DDPG} &                 \makecell{SAC} &            \makecell{PPS} \\
\midrule
    1D Double Integrator &  \makecell{$22.5\% $\\$(\pm 2.4\%)$} &  \makecell{$18.9\% $\\$(\pm 4.8\%)$} &  \makecell{$\bm{93.7\% }$\\$(\pm 1.0\%)$} \\
    2-Link Planar Arm &  \makecell{$32.1\% $\\$(\pm 3.0\%)$} &   \makecell{$9.3\% $\\$(\pm 2.0\%)$} &  \makecell{$\bm{84.6\% }$\\$(\pm 0.5\%)$} \\
              Acrobot &   \makecell{$0.1\% $\\$(\pm 0.5\%)$} &   \makecell{$1.8\% $\\$(\pm 3.4\%)$} &  \makecell{$\bm{15.0\% }$\\$(\pm 0.3\%)$} \\
         Mountain Car &  \makecell{$52.5\% $\\$(\pm 8.9\%)$} &  \makecell{$16.1\% $\\$(\pm 3.9\%)$} &  \makecell{$\bm{54.4\% }$\\$(\pm 1.5\%)$} \\
\bottomrule
\end{tabularx}
\caption{State-space coverage as visited percentage of bins. Each number represents the median
  of ten independent runs, the interquartile range is listed in brackets.
}
\label{tbl:coverage}
\end{table}

The exploration as measured by the proportion of state space covered is
shown in \reftbl{tbl:coverage}. The median coverage of ten independent
runs as well as the interquartile range (IQR; in parentheses)
after $10^5$ interaction steps are listed. In all four
environments \gls{PPS} achieves the highest state-space coverage, with
a margin significantly larger than the IQR in all cases but the
Mountain Car. DDPG appears to be exploring more than SAC in most
cases, except when exploration is very low (Acrobot), where both SAC
\& DDPG explore very little.

\subsection{\ref{q_stuck} -- Susceptibility to local optima}\label{a_stuck}
\reffig{fig:learning_curves} depicts the evaluation returns during
learning for the baselines (\gls{DDPG}, \gls{SAC}), as well as
\gls{PPS}. In each environment ten independent runs are performed.
The median evaluation return is depicted, with the shaded area
showing the IQR around the median.
We use tanh activations in the indirectly-trained policies (and PPS)
because we found them to produce results with smaller variance and to
perform more robustly when learning from the fixed buffer.

While the collection phase consists of $10^5$ steps, the plots depict
learning curves for $2\cdot 10^5$ steps. In the indirect cases, the
replay buffer is fixed so the $x$-axis steps only correspond to
learning steps, not to environment interaction steps -- which are mixed
(collecting \& learning) in the case of the direct algorithms \gls{SAC}
and \gls{DDPG}.

The results show improved performance when training on the fixed
replay buffer. They also show performance of \gls{PPS} to be superior
to the policy indirectly trained on \gls{SAC} data as well as the
directly-trained \gls{SAC} policy. However, from the learning curves,
it appears the variance is higher in the \gls{PPS} learning process,
and in some cases (Mountain Car, Double Integrator) dips after initially
rising, before reaching high performance.
On the Acrobot \gls{PPS}
performs a lot better than the other algorithms. The Acrobot is
a hard problem because it is underactuated and taking the wrong
actions will push the agent back to a state that is further away from
the goal. Moreover, the reward is very sparse, with zero reward except
for reaching the goal region. Here, the directed exploration of
\gls{PPS} is able to unleash its full potential.

In the Mountain Car environment, \gls{PPS} reaches a similar
performance to training SAC from DDPG data, while the policies trained
by DDPG exhibit high variance in their returns and SAC achieves zero
returns. This striking result of SAC presumably stems from the reward
structure in the Mountain Car environment, where a large positive
reward is attributed when the goal is reached, while each action is
penalized according to its effort. Here, SAC optimizes for the closest
local minimum: not taking any actions. In the cases of the Double Integrator and
the Planar Arm, both DDPG and SAC pick up on the shaped reward and
optimize towards the local minimum, while PPS, although with less data
density and thus higher variance, reaches the high-reward areas.
Surprisingly in many cases, it appears that \gls{SAC} is more
succesful when learning from \gls{DDPG} data than when learning from
\gls{SAC} data itself.

\section{Discussion}\label{sec:discussion}

In this work, we investigated whether directed exploration in
\gls{PPS} reduces the probability of getting stuck in a local optimum
compared to undirectedly exploring gradient based \gls{D-RL}
methods. Our experiments show that this can indeed be a problem in
practice, even in relatively small toy problems such as the double
integrator with two local optima. On average the agent controlled by
\gls{PPS} explores a wider part of the state space than \gls{D-RL}
methods that focus on reward accumulation.  While this is partly
expected due to the exploration/exploitation tradeoff, our
experiments show that even from this highly-exploratory data good
policies can be learned and can even surpass the performance of the
agents focusing more on exploitation. This shows that by using
directed exploration \gls{PPS} reduces the probability of getting
stuck in local optima compared to undirectedly exploring \gls{D-RL}
methods.

The data gathered by \gls{PPS} are not biased by reward accumulation
and are thus more representative of the environment. This should make
these data quite suitable for reuse in different tasks. This is
also visible in \reftbl{tbl:coverage} where the coverage is reduced
for the underactuated and thus harder-to-control environments
(Acrobot, Mountain Car) where the coverage is impacted by the
difficulty of the environment dynamics. Also in these cases the
coverage achieved by \gls{PPS} dominates that of the baseline
algorithms.

Although \gls{PPS} is able to achieve better policies than the other methods, it
also exhibits more variance in the learning process. Presumably this
stems from the extreme off-policy nature of our algorithm which should
increase the variance in the gradient estimates. This instability during
continued learning on the fixed buffer (timesteps beyond
$10^5$ in \reffig{fig:learning_curves}) is also present
when learning from data collected by the \gls{D-RL} methods and
warrants future research.

One limitation of our study is that the evaluations are done on
relatively low-dimensional tasks. Uniformly covering the state space
might reach its limitations in high-dimensional problems due to the
curse of dimensionality: A goal or high-reward region might be too
narrow to be covered by uniformly-distributed samples. However, this
problem can be alleviated by changing the uniform sampling mechanism
to a reward-adaptive procedure which we will investigate in future work.
We envision our method to be most useful in scenarios where a model is
already available (for example model-based RL with an already-learned
model to adapt to a different task) or even more importantly in the
regime of Sim2Real training where a simulation model is used to train
robust policies that can then be applied to the real system. In recent
years the approach of domain randomization has gained attention where
properties of the simulation model are randomized thereby forcing the
policy to be robust to model inaccuracies and preventing overfitting
to idiosyncracies of the simulation. In this setting our method has
the potential to speed up domain-randomized training: By randomizing
the model, using planning to quickly discover potential new goal
regions and adapting the sampling to re-use prior knowledge of similar
tasks, the method can potentially focus the training on relevant parts
of the state space and reduce the number of necessary samples.  This
will be evaluated in future work.
 
\begin{algorithm}[!bt]
  \begin{algorithmic}
    \Function{$\text{AQR}_\text{distance}$}{$x_0, x_r$} \State $\text{Let } A, B, c$ \Comment{linearized dynamics around $x_r, x_0$}
    \State $ \bar{x}_0 = x_0 - x_r$
    \State $\dot P(t) = A P(t) + P(t) A^T - B R^{-1} B ^T,$
    \State $~~~ P(T) = 0, t\in\{T, ..., 0\}$
    \State $S^{-1}(t) = P(t)$
    \State $d(\overline x_0, T) = e^{A T} \overline x_0 + \int_0^T e^{A(T-\tau)} c ~ d \tau$
    \State $J^*(\overline x, T) = T + \frac{1}{2} d^\top(\overline x_0, T) S^{-1}(T)d(\overline x_
    0, T)$
    \State $ T^* = \argmin_T J^*(\bar{x}_0, T), ~~ 0 \leq T \leq T_{max} $
    \State \Return $ J^*(\bar{x}_0, T^*), T^*$
    \EndFunction
    \Function{$\text{AQR}_\text{Nearest}$}{$\mathcal{T}, x_r$}
    \State \Return $ \argmin_{v \in \mathcal{T}} $  \Call{$\text{AQR}_\text{distance}$}{$x_v, x_r$}
    \EndFunction
  \end{algorithmic}
  \caption{Calculate AQR \& find nearest neighbor}
  \label{alg:aqr}
\end{algorithm}

\begin{algorithm}[!bt]
  \begin{algorithmic}
    \Function{Steer}{$\vclosest, \xtarget, h, G^\affnzd, H^\affnzd$}
    \State $d \gets ()$ \Comment{empty trajectory}
    \State $s \gets \vclosest$ \Comment{initalize state}
    \State $\xtarget \overset{\affnzd}{\gets} \xtarget$
    \For{$L \in \{h, \ldots, 1\}$} \Comment{Shrinking Horizon}
    \State $x_0 \overset{\affnzd}{\gets} s$ \Comment{all $x$ are affinized}
    \State $
    \begin{array}{ll}
      \mbox{With}       & Q_L \gg Q \\
      \mbox{minimize}   &  (x_L-x_r)^T Q_L (x_L-x_r) + \\
                        & \sum_{k=0}^{L-1} (x_k-x_r)^T Q (x_k-x_r) + \\
                        & \sum_{k=0}^{L-1} a_k^T R a_k \\
      \mbox{subject to} & x_{k+1} = G^\affnzd x_k + H^\affnzd a_k \\
                        & x_{\rm min} \le x_k  \le x_{\rm max} \\
                        & a_{\rm min} \le a_k  \le a_{\rm max} \\
\end{array}$
\State $s', r \gets \text{step}(a_0)$ \Comment{Take environment step}
    \State $d \gets d \oplus (s, a_0, r, s')$ \Comment{add to trajectory}
    \State $s \gets s'$
    \EndFor
    \State \Return $s, d$ \Comment{reached state, trajectory}

    \EndFunction
  \end{algorithmic}
  \caption{Local steering model-predictive control method.}
  \label{alg:steer}
\end{algorithm}

\begin{table}[!tb]
\caption{Description of the 1D double-integrator test environment: a point mass $M$ can be moved in a
    one-dimensional space $X=(\text{position }, \text{velocity})$ by
    applying a continuous-valued force. Reward is received based on
    the distance to two possible goal locations ($G_1, G_2$).
  }
\renewcommand{\arraystretch}{1.3}
\resizebox{\columnwidth}{!}{
     \begin{tabular}{@{}l@{\hskip 3em}l@{\hskip 3em}l@{\hskip 3em}l@{\hskip 3em}l@{}}
     \toprule
Dynamics \\
      $X = \begin{bmatrix}x \\ \dot x \end{bmatrix}$ &
      $\dot x = Ax + Bu$ &
                           \multicolumn{2}{c}{\multirow{3}{*}{  \resizebox{.3\textwidth}{!}{
                           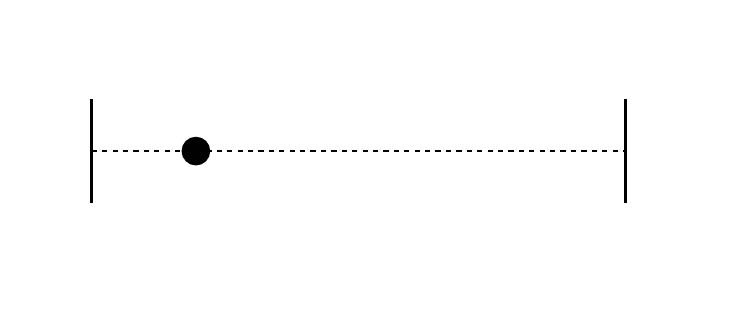}}} \\[3ex]
      $  A = \begin{bmatrix}
         0 & 1 \\
         0 & 0
       \end{bmatrix}$
       &
      $ B = \begin{bmatrix}
       0 \\
       1
         \end{bmatrix}$ &    \multicolumn{2}{c}{} \\
       \midrule
       Reward \\
       \multicolumn{2}{@{}l}{$\max((1-\tanh |X-G^*_1|),$ $2(1-\tanh |X-G^*_2|))$}  &
       $G_1 = \begin{bmatrix} -2.5 \\ 0.0 \end{bmatrix}$ &
       $G_2 = \begin{bmatrix}  6.0 \\ 0.0 \end{bmatrix}$ \\[3ex]
       \midrule
       Limits \\
       $ u \in [-1; 1]$ & $x \in [ -10; 10 ]$ &  $\dot x \in [ -2.5; 2.5] $ \\
     \bottomrule
     \end{tabular}
   }
   \label{tbl:envdescription}
 \end{table}

\section{Appendix}\label{app:aqr_metric}
The approach implemented to calculate the AQR distance and to find the nearest state to a goal state $x_r$ is summarized in Algorithm~\ref{alg:aqr}.
The steering approach based on a linear MPC with shrinking horizon is summarized in Algorithm~\ref{alg:steer}.
 
\bibliography{rlvsplanning-ws.autogen.bib,additional-bibliography.bib}

\end{document}